\documentclass[10pt,twocolumn,conference]{IEEEtran}
\usepackage{times}
\usepackage{graphics}
\usepackage{epstopdf}
\usepackage[reqno]{amsmath}
\usepackage{amsfonts}
\usepackage{caption2}
\usepackage{times,amsmath,epsfig}
\usepackage{latexsym,amssymb}
\usepackage{rotating,color}
\usepackage{cite}
\usepackage{extarrows}
\usepackage{psfrag}
\usepackage{float,stfloats,multirow,subfigure}
\usepackage[]{algorithm2e}
\psfull

\newcommand{\lettrineo}[2]{%
\settowidth{\lwidth}{#2\kern2pt}%
\noindent\hangindent\lwidth\hangafter-#1\hskip-\lwidth%
\smash{\hbox to\lwidth{\raise7pt\vtop{\null\hbox{#2}}%
\hfill}}\ignorespaces}

\newfont{\HUGEfonto}{cmr17 scaled \magstep3}  %  cmbx12
\DeclareSymbolFont{AMSa}{U}{msa}{m}{n}
\DeclareMathSymbol{\blacksquare}  {\mathord}{AMSa}{"04}

\newfont{\bbb}{msbm10 scaled 500}

\newfont{\bb}{msbm10 scaled 1100}

\newcommand{\thetav}{\boldsymbol{\theta}}

\renewcommand{\det}{{\hbox{det}}}

\renewcommand{\arg}{{\hbox{argmax}}}

\newcommand{\beq}{\begin{equation}}
\newcommand{\enq}{\end{equation}}
\newcommand{\beqa}{\begin{eqnarray}}
\newcommand{\enqa}{\end{eqnarray}}

\newcommand{\beql}[1]{\begin{equation}\label{#1}}

\newcommand{\be}{\beta}

\newcommand{\qed}{\hfill $\Box$}

\newtheorem{thm}{Theorem}

\def\bbC{{\sf C}\kern -6pt {\sf C}}
\def\bbF{{\sf F}\kern -5pt {\sf F}}
\def\bbR{{\sf R}\kern -6pt {\sf R}}
\def\bbZ{{\sf Z}\kern -5pt {\sf Z}}
\def\sfbegin{\begingroup\sf}
\def\sfend{\endgroup}

\def\be{\begin{eqnarray*}}
\def\ee{\end{eqnarray*}}

\newcommand{\no}{\nonumber}

\DeclareMathOperator{\tr}{trace}

\IEEEoverridecommandlockouts
\begin{document}
\title{Quick Best Action Identification in Linear Bandit Problems}
\author{Jun Geng$^{1}$ and Lifeng Lai$^{2}$\\
	$^{1}$ School of Electrical and Information Engineering, Harbin Institue of Technology, Harbin, China \\
	$^{2}$ Dept. of Electrical and Computer Engineering, University of California, Davis, CA\\ Emails: jgeng@hit.edu.cn, lflai@ucdavis.edu
	\thanks{
		The work of J. Geng is supported by the National Natural Science Foundation of China under grant 61601144. The work of L. Lai is supported by the National Science Foundation under grants ECCS-1711468 and CCF-1717943.}}

%\date{}
\maketitle %\pagestyle{plain}

%%%%%%%%%%%%%%% Article Body %%%%%%%%%%%%%%%%%%%%%%%%%%%%%%%%%%%%%%%%%
%%%%=======================================================

\begin{abstract}
In this paper, we consider a best action identification problem in the stochastic linear bandit setup with a fixed confident constraint. %In the stochastic linear bandit setup, at each time slot $t$, a learner chooses an action $\mathbf{x}_{t}$ from a feasible set ${D}$ and obtains a reward $y_{t} = <\mathbf{x}_{t}, \boldsymbol{\theta}^{*}>+\eta_{t}$, in which $\boldsymbol{\theta}^{*}$ is a fixed but unknown parameter and $\eta_{t}$ is the noise.
In the considered best action identification problem, instead of minimizing the accumulative regret as done in existing works, the learner aims to obtain an accurate estimate of the underlying parameter based on his action and reward sequences. To improve the estimation efficiency, the learner is allowed to select his action based his historical information; hence the whole procedure is designed in a sequential adaptive manner. We first show that the existing algorithms designed to minimize the accumulative regret is not a consisent estimator and hence is not a good policy for our problem. We then charcaterize a lower bound on the estimation error for any policy. We further design a simple policy and show that the estimation error of the designed policy achieves the same scaling order as that of the derived lower bound. % In this paper, we propose an action selection rule and a parameter estimation method based on the convex optimization, hence the proposed method can be computed efficiently. We also provide an upper bound for the estimation error, which further indicates that the estimate $\hat{\boldsymbol{\theta}}$ converges to the true underlying $\boldsymbol{\theta}^{*}$ in probability as the number of action round goes to infinity.
\end{abstract}

%\begin{keywords}
%stochastic linear bandit, fixed confidence
%\end{keywords}

%%%%%%%%%%%%%%% Article Body %%%%%%%%%%%%%%%%%%%%%%%%%%%%%%%%%%%%%%%%%
%%%%=======================================================

\section{Introduction} \label{sec:intro}
Multi-armed bandit problem is a canonical sequential decision problem that has a wide range of applications~\cite{Press:NAS:09,Awerbuch:STOC:04,Manickam:ICASSP:17, Reverdy:ICASSP:18,Lai:TMC:11}. In the classic multi-armed bandit problem, at each time slot, a decision maker has to choose one of $K$ competing decisions or ``arms'', and receives a reward related to certain unknown parameters from his selected decision. %The rewards of each decision is usually a random variable; hence it costs the decision maker several rounds if he wants to infer the expected rewards of a decision.
Based on the knowledge collected from his past decisions and the corresponding rewards, the decision maker can then carefully decide his future actions according to different goals. The most commonly used goal is to minimize the cumulative regret, which is the cumulative difference between the optimal reward that one can achieve when the underlying parameters are known and the reward of the action taken by the decision maker. This setup nicely captures ``exploration versus exploitation'' phenomena in sequential decision making, as a crucial tradeoff faced by the decision maker at each round is between ``exploitation'', i.e. to choose the decision with the highest estimated expected rewards, and ``exploration'', i.e. to choose other decisions so as to obtain better estimates of the expected rewards of these decisions. Recently, another goal named ``best arm identification" has received significant attentions~\cite{Gabillon:NIPS:12, Jamieson:CISS:14, Jamieson:JMLR:14,Karnin:ICML:13,Jamieson:arXiv:13,Auer:JMLR:02, Kalyanakrishnan:ICML:12}. In the best arm identification problem, instead of minimizing the cumulative regret, the goal is to identify the best arm that provides the highest expected rewards with high probability. This setup is also known as pure exploration since the decision maker now has the freedom to explore all arms without having to worry about regrets incurred in these exploration actions.

%Multi-armed bandit problem has a wide range of potential applications. For example, in clinical trials \cite{Press:NAS:09}, each decision is to choose one of $K$ experimental treatment methods, and the goal is to minimize the loss -- the damage on the health -- of the patient. In the problem of adaptive routing \cite{Awerbuch:STOC:04}, it is of interest to send packages from source to destination via multiple routers. The delay of each route is unknown. Each decision is to select one of $K$ routes and the goal is to minimize the total network delay. Multi-armed bandit problem is also closely related to the reinforcement learning problem \cite{Manickam:ICASSP:17, Reverdy:ICASSP:18} in machine learning area, which receives increasing attention in recent years.

A natural generalization of the classic multi-armed bandit problem is so called stochastic linear multi-armed bandit problem \cite{Dani:COLT:08}. In the stochastic linear multi-armed bandit problem, the decision maker chooses his decision $\mathbf{x}_{t}$ from an $d-$dimensional compact set ${D}$ and receives a reward $<\mathbf{x}_t, \boldsymbol{\theta}^{*}>+\eta_{t}$, in which $\boldsymbol{\theta}^{*}$ is a fixed but unknown parameter and $\eta_{t}$ is noise. Defining the regret as the difference between the rewards of the best decisions when $\boldsymbol{\theta}^{*}$ is known and the rewards of the selected decisions, existing works on the stochastic linear multi-armed bandit problem aim to minimize the total regret. For example, \cite{Dani:COLT:08, Abbasi:NIPS:11} have proposed algorithms according to the optimism in the face of uncertainty (OFU) principle, and have shown the proposed algorithms are Hannan consistent.

In this paper, similar to the best arm identification problem studied in the classic multi-armed bandit setup, we consider the best action identification problem in the stochastic linear multi-armed bandit setup. More specifically, instead of aiming to minimize the cumulative regret, we aim to obtain an accurate estimation $\hat{\thetav}$ of the unknown parameter $\boldsymbol{\theta}^{*}$ under a fixed confidence constraint. In particular, the decision maker aims to minimize the total number of actions under the constraint that the estimation error $|| \hat{\thetav} - \boldsymbol{\theta}^{*}||_2$ is under control with a large probability. We call this best action identification problem, as the best action $\mathbf{x}_t$ should have the same direction as $\boldsymbol{\theta}^{*}$.

In this paper, we first show that existing algorithms based on the OFU principle lead to inconsistent estimators of $\boldsymbol{\theta}^{*}$ and hence are not suitable for the best action identification. Intuitively, the OFU algorithm keeps selecting the actions that are close to the current estimation $\hat{\boldsymbol{\theta}}_{t}$ in each round since it aims to minimize the regret. As a result, all selected actions are concentrated in a small cone around the direction of the true underlying parameter $\boldsymbol{\theta}^{*}$. The decision maker has to use the rewards of selected actions to estimate $\boldsymbol{\theta}^{*}$, but the actions with similar directions only bring similar rewards. In other words, it is challenging for the decision maker to tell whether the change of rewards is caused by the different action selection or by the random noise. Hence it is difficult to identify which action is better. Motivated by this intuitive explanation, we propose a scheme that selects actions that are orthogonal to the direction of $\hat{\boldsymbol{\theta}}_{t}$. We show that the rewards from these different directions are effective in identifying the best action. In particular, we show that the proposed algorithm leads to a consistent estimator of $\boldsymbol{\theta}^{*}$. Furthermore, we calculate a lower bound of the estimation error of any policy, and further show that the performance of our proposed algorithm achieves this lower bound up to a constant factor.

%The fixed confidence constraint has been considered in several existing works on the classic multi-arm bandit problem \cite{Gabillon:NIPS:12, Jamieson:CISS:14, Jamieson:JMLR:14}. Several effective algorithms, including successive elimination \cite{Karnin:ICML:13}, PRISM \cite{Jamieson:arXiv:13}, upper confidence bound (UCB)\cite{Auer:JMLR:02, Kalyanakrishnan:ICML:12}, etc. have been proposed and analyzed. These works aim to identify the choice or the ``arm'' providing the highest expected rewards with high probability while keeping the number of samples minimized. However, in our work, we are interested in estimating the unknown parameter $\boldsymbol{\theta}^{*}$. Our constraint is to keep the estimation error under control in a high probability, and each decision in our context is a $d-$dimensional vector.

The remainder of the paper is organized as follows. The mathematical model is given in Section \ref{sec:model}. In Section \ref{sec:alg}, the limitation of OFU based algorithms is discussed. We further propose a new algorithm and analyze its performance. Section \ref{sec:sim} provides a numerical example to illustrate the conclusion obtained in this paper.  Section \ref{sec:con} offers some concluding remarks. %Due to space limitations, we present only main ideas and conclusions. Details of proofs can be found in~\cite{Jun:TIT:14}.

{\bf Notations}: $||\mathbf{x}||_{p}$ denotes the $p-$norm of a vector $\mathbf{x} \in \mathbb{R}^{d}$. For a positive definite matrix $\mathbf{A}\in \mathbb{R}^{d \times d}$, the weighted norm of a vector $\mathbf{x}$ is denoted as $||\mathbf{x}||_{\mathbf{A}}=\sqrt{\mathbf{x}^{T}\mathbf{A}\mathbf{x}}$, and the weighted inner product of two vectors $\mathbf{x}, \mathbf{y}$ is denoted as $<\mathbf{x}, \mathbf{y}>_{\mathbf{A}} = \mathbf{x}^{T}\mathbf{A}\mathbf{y}$. $\lambda_{\max}(\mathbf{A})$, $\lambda_{\min}(\mathbf{A})$, $\det(\mathbf{A})$ and $\tr(\mathbf{A})$ denote the maximum eigenvalue, the minimum eigenvalue, the determinant and the trace of matrix $\mathbf{A}$, respectively.

\section{Problem Formulation} \label{sec:model}
In this paper, we consider the stochastic linear bandit problem which proceeds in rounds $t=1, 2, \ldots$. In each round $t$, the decision maker chooses a decision $\mathbf{x}_{t}$ from a compact decision set $D_{t} \subset \mathbb{R}^{d}$, and subsequently obtains a reward
\begin{eqnarray}
y_{t} = <\mathbf{x}_{t}, \boldsymbol{\theta}^{*}> + \eta_{t},
\end{eqnarray}
in which $\boldsymbol{\theta}^{*} \in \mathbb{R}^{d}$ is a fixed but unknown parameter with finite $l_{2}$-norm $||\boldsymbol{\theta}^{*}||_{2}\leq S$, and $\eta_{t}$ is a centered sub-Gaussian random variable with variance proxy $\sigma^{2}$. $\{ \eta_{t}, t=1, 2, \ldots \}$ is assumed to be a sequence of independently and identically distributed (i.i.d.) random variables.

Let $\mathcal{F}_{t} = \sigma\{\eta_{1}, \eta_{2}, \ldots, \eta_{t}\}$ be the sigma field at time $t$. The decision maker is allowed to choose his decision adaptively based on his historical information. Mathematically, $\mathbf{x}_{t}$ can be expressed as
\begin{eqnarray}
\mathbf{x}_{t} = f_{t}(\mathbf{x}_{1}, y_{1}, \ldots, \mathbf{x}_{t-1}, y_{t-1}), \no
\end{eqnarray}
in which $f_{t}(\cdot)$ is some $\mathcal{F}_{t-1}$ measurable function. To simplify the derivation, we assume that the decision set $D_{t}=\{\mathbf{x} \in \mathbb{R}^{d}: ||\mathbf{x}||_{2}^{2} \leq 1\}$, which is a fixed set over time. Hence in the remainder of this paper, we also denote the decision set as $D$.

We express the relationship between decisions and corresponding rewards in the matrix form as
\begin{eqnarray}
\mathbf{Y}_{t} = \mathbf{X}_{t} \boldsymbol{\theta}^{*} + \boldsymbol{\eta}_{t} \label{eq:linear_model},
\end{eqnarray}
in which $\mathbf{Y}_{t} = [y_{1}, y_{2}, \ldots, y_{t}]^{T}$, $\boldsymbol{\eta}_{t} = [\eta_{1}, \eta_{2}, \ldots, \eta_{t}]^{T}$ and $\mathbf{X}_{t} = [\mathbf{x}_{1}^{T}, \mathbf{x}_{2}^{T}, \ldots, \mathbf{x}_{t}^{T}]^{T}  \in \mathbb{R}^{t\times d}$.
Denote $\hat{\boldsymbol{\theta}}_{t}$ as the estimate of $\boldsymbol{\theta}^{*}$ at time $t$. The decision maker aims to design an efficient algorithm to select decisions $\mathbf{X}_{t}$ and accurately estimate the unknown parameter $\boldsymbol{\theta}^{*}$ based on his sequential information $\{\mathbf{x}_{1}, \ldots, \mathbf{x}_{t}, y_{1}, \ldots, y_{t}\}$. The performance metric is specified as
\begin{eqnarray}
P(||\hat{\boldsymbol{\theta}}_{t} - \boldsymbol{\theta}^{*}||_{2}^{2} \leq \epsilon ) \geq 1-\delta \label{eq:err_prob}
\end{eqnarray}
for some given constant $\epsilon > 0$ and $\delta \in (0, 1)$. That is, the decision maker should have strong confidence on the result that the estimation error is less than a small value $\epsilon$ when the decision procedure is terminated. Since $\{\eta_{t}\}$ is a sequence of sub-Gaussian random variable, we expect that $\epsilon$ converges to zero and $\delta$ decays exponentially with respect to $t$ as $t \rightarrow \infty$.

\section{Algorithms and Performances} \label{sec:alg}
A natural estimator for \eqref{eq:linear_model} is the ordinary least squares estimator
\begin{eqnarray}
\hat{\boldsymbol{\theta}}_{t} = (\mathbf{X}_{t}^{T}\mathbf{X}_{t})^{-1}\mathbf{X}_{t}^{T}\mathbf{Y}_{t}.\label{eq:mmse}
\end{eqnarray}

One difficulty with the above estimator is that $\mathbf{X}_{t}^{T}\mathbf{X}_{t}$ is not invertible when its rank is deficient (e.g. $t \leq d$). In this paper we focus on the following class of estimators that are slight modification of~\eqref{eq:mmse}
\begin{eqnarray}
\hat{\boldsymbol{\theta}}_{t} = (\mathbf{X}_{t}^{T}\mathbf{X}_{t} + \mathbf{W}_{0})^{-1}\mathbf{X}_{t}^{T}\mathbf{Y}_{t}, \label{eq:estimator}
\end{eqnarray}
in which $\mathbf{W}_{0}$ is a positive definite matrix. This class of estimators are widely used in the regret minimizaiton problems~\cite{Dani:COLT:08, Abbasi:NIPS:11}. For notation convenience, we define
\begin{eqnarray}
\mathbf{W}_{t} := \mathbf{W}_{0} + \mathbf{X}_{t}^{T}\mathbf{X}_{t} . \label{eq:WtW0}
\end{eqnarray}
It is easy to see that $\mathbf{W}_{t}$ is always positive definite; hence the inversion in \eqref{eq:estimator} is always valid. We further note that $\mathbf{W}_{t}$ can be efficiently calculated using the recursive formula $\mathbf{W}_{t} = \mathbf{W}_{t-1} + \mathbf{x}_{t}\mathbf{x}_{t}^{T}$.

\subsection{Challenges of Existing Algorithms}
The most well known algorithm for the stochastic linear bandit problem is designed according to the \emph{optimism in the face of uncertainty principle} \cite{Abbasi:NIPS:11}. The basic idea of this algorithm is to use observations to construct a confidence set $C_{t} \subset \mathbb{R}^{d}$ that contains the unknown parameter $\boldsymbol{\theta}^{*}$ with a high probability. The confidence set $C_{t}$ is updated whenever the decision maker obtains a new reward $y_{t}$. The algorithm then estimates the unknown parameter by $\hat{\boldsymbol{\theta}}_{t} = \arg_{\boldsymbol{\theta}\in C_{t}}(\max_{\mathbf{x}\in D_{t}}<\mathbf{x}, \boldsymbol{\theta}>)$ and selects the next decision by solving $\mathbf{x}_{t} =\arg_{\mathbf{x}\in D_{t}}<\mathbf{x}, \hat{\boldsymbol{\theta}}_{t}>$.

In our context, for $t = 1, 2, \ldots$, the algorithm designed according to the OFU principle can be expressed as:
\begin{eqnarray}
&&\hat{\boldsymbol{\theta}}_{t} = (\mathbf{X}_{t}^{T}\mathbf{X}_{t} + \mathbf{W}_{0})^{-1}\mathbf{X}_{t}^{T}\mathbf{Y}_{t}, \label{eq:the_update}\\
&&C_{t} = \left\{ \boldsymbol{\theta} \in \mathbb{R}^{d}: ||\hat{\boldsymbol{\theta}}_{t} - \boldsymbol{\theta}||_{\mathbf{W}_{t}} \leq \beta_{t} \right\}, \label{eq:c_update} \\
&&\mathbf{x}_{t} = \text{argmax}_{\mathbf{x}\in D_{t}} <\mathbf{x}, \hat{\boldsymbol{\theta}}_{t}> = \hat{\boldsymbol{\theta}}_{t}/||\hat{\boldsymbol{\theta}}_{t}||_{2}^{2}. \label{eq:x_update}
\end{eqnarray}
We note that the confidence region $C_{t}$ is an ellipsoid with radius $\beta_{t}$. The value of $\beta_{t}$ is updated at every time slot according to newly obtained information.

Several existing works \cite{Dani:COLT:08, Abbasi:NIPS:11} have shown that, if $\beta_{t}$ is properly designed, the above algorithm has a small cumulative regret. Particularly, let $\mathbf{x}_{t}^{*} = \arg_{\mathbf{x} \in D_{t}} <\mathbf{x}, \boldsymbol{\theta}^{*}>$ be the best decision for $\boldsymbol{\theta}^{*}$, let $r_{t} = <\mathbf{x}_{t}^{*}, \boldsymbol{\theta}^{*}> - <\mathbf{x}_{t}, \boldsymbol{\theta}^{*}>$ be the regret at time $t$ for taking decision $\mathbf{x}_{t}$ and let $R_{n} = \sum_{t=1}^{n} r_{t}$ be the cumulative regret. \cite{Abbasi:NIPS:11} proved the following result.
\begin{thm} \label{thm:OFU_regret}
\emph{(Theorem 2 and Theorem 3 in \cite{Abbasi:NIPS:11})} Let $\mathbf{W}_{0} = \kappa\mathbf{I}$, $\kappa >0$. By setting
$$\beta_{t} = \sigma^2\sqrt{2\log(\det(\mathbf{W}_{t})^{1/2}\det(\lambda\mathbf{I})^{-1/2}/\delta)}+\kappa^{1/2}S, $$
then for any $\delta >0$, with probability at least $1-\delta$, $\boldsymbol{\theta^{*}}$ lies in the set $C_{t}$. Further more, if for all $t$ and all $\mathbf{x}\in D_{t}$, $<\mathbf{x}, \boldsymbol{\theta^{*}}> \in [-1, 1]$, then with probability at least $1-\delta$, the cumulative regret satisfies
\begin{eqnarray}
&&\forall n\geq 0, \quad R_{n} \leq 4\sqrt{nd\log(\kappa+n/d)}\no\\
&&\hspace{10mm}(\kappa^{1/2}S + \sigma^2\sqrt{2\log(1/\delta)+d\log(1+n/(\kappa d))}). \no
\end{eqnarray}
\end{thm}

Theorem~\ref{thm:OFU_regret} indicates that the OFU algorithm is Hannan consistent, i.e., $\lim_{n \rightarrow \infty} R_{n}/n = 0$. However, in the following, we point out that the OFU algorithm leads to an inconsistent estimate of $\boldsymbol{\theta^{*}}$. The result is stated in the following theorem.
\begin{thm} \label{thm:OFU_error}
If $R_{n}/n \rightarrow 0 \text{ as } n\rightarrow \infty$ with probability at least $1-\delta$, then
\begin{eqnarray}
\lim_{t\rightarrow\infty} P\left( ||\hat{\boldsymbol{\theta}}_{t} - \boldsymbol{\theta^{*}}||_{2}^2 \geq \sigma^2 \right) \geq 1-\delta.
\end{eqnarray}
\end{thm}
\begin{IEEEproof}
Please see Appendix \ref{adp:OFU_error}.
\end{IEEEproof}

Define event $\mathcal{E}:= \left\{ ||\hat{\boldsymbol{\theta}}_{t} - \boldsymbol{\theta^{*}}||_{2}^2 \geq \sigma^2 \right\}$. Theorem \ref{thm:OFU_error} implies that
\begin{eqnarray}
\mathbb{E}[||\hat{\boldsymbol{\theta}}_{t} - \boldsymbol{\theta^{*}}||_{2}^2] \geq \mathbb{E}[||\hat{\boldsymbol{\theta}}_{t} - \boldsymbol{\theta^{*}}||_{2}^2 | \mathcal{E}] P(\mathcal{E}) \geq \sigma^2(1-\delta). \no
\end{eqnarray}
That is, the estimation error of the OFU algorithm does not vanish as the sample size goes to infinity. %\emph{The estimation error is a constant, which is not decay with the increasing number of observations}.

\begin{figure}[h]
\begin{center}
\includegraphics[width=9cm]{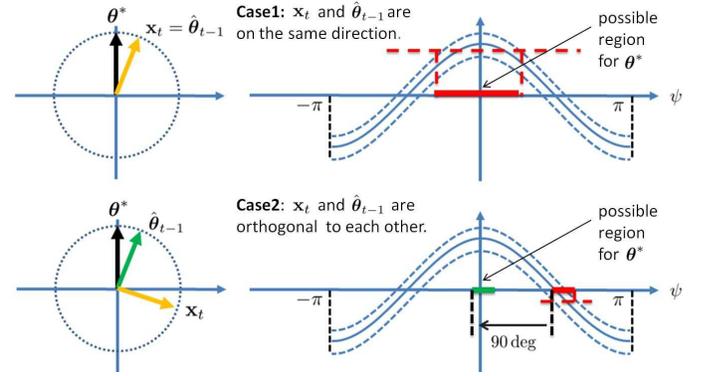}
\caption{An illustration of the difference between the OFU algorithm and the proposed algorithm. Upper figures: the case for OFU algorithm. Lower figures: the case for proposed algorithm.}
\label{fig:intution}
\end{center}
\end{figure}

In the following, we provide an intuitive explanation of the reason why OFU algorithms lead to inconsistent estimators. The examination of this also provides motivation for the proposed algorithm to be discussed below. Considering the case with $d=2$, the reward $y_{t} = <\mathbf{x}_{t}, \boldsymbol{\theta}^{*}> + \eta_{t} = ||\mathbf{x}_{t}||||\boldsymbol{\theta}^{*}||\cos\psi + \eta_{t}$, in which $\psi$ is the angle between $\mathbf{x}_{t}$ and $\boldsymbol{\theta}^{*}$. As illustrated in the upper-right subfigure in Fig. \ref{fig:intution}, the solid cosine curve is $<\mathbf{x}_{t}, \boldsymbol{\theta}^{*}>$, and the region bounded by the two dash cosine lines characterizes the possible region of the rewards $y_{t}$. In OFU algorithms, the decision maker takes action $\mathbf{x}_{t} = \hat{\boldsymbol{\theta}}_{t-1}$. As $\hat{\boldsymbol{\theta}}_{t-1}$ and $\boldsymbol{\theta}^{*}$ are generally close, the regret is small and $\psi$ is close to zero. In this case, an obtained feedback reward $y_{t}$ leads to a wide possible range for $\boldsymbol{\theta}^{*}$. That is, any value of $\psi$ in the red region of the upper-right subfigure in Fig. \ref{fig:intution} could lead to the same reward $y_{t}$. In the regret minimization, this is unavoidable, as we need to select $\mathbf{x}_{t}$ that has small angle with $\boldsymbol{\theta}^{*}$. In our problem setup, as the regret is not of primary concern, we can avoid this by selecting $\mathbf{x}_{t}$ to be orthogonal to $\hat{\boldsymbol{\theta}}_{t-1}$ (and hence has large angle with $\boldsymbol{\theta}^{*}$). These actions are helpful in improving the estimation accuracy of $\psi$ as their rewards are close to the zero-crossing region of the cosine curve, which infers a much narrower possible region for $\boldsymbol{\theta}^{*}$. The proposed algorithm to be discussed below is motivated by this observation.

\subsection{Proposed Algorithm and Performance Analysis}
%In the rest of this paper, we focus our discussion on the estimation strategy
%

Motivated by the discussion above, we propose a novel algorithm which leads to a consistent estimator with a fast convergence rate. The proposed algorithm is specified in Algorithm \ref{alg:proposed}. To facilitate the presentation, for $k=1, 2, \ldots$, we use the following notations in Algorithm \ref{alg:proposed}:
\begin{eqnarray}
&&\mathbf{X}_{k,d} = \left[ \mathbf{x}_{(k-1)d+1}^{T}, \mathbf{x}_{(k-1)d+2}^{T}, \cdots, \mathbf{x}_{kd}^{T} \right]^{T}, \no\\
&&\mathbf{Y}_{k,d} = [y_{(k-1)d+1}, y_{(k-1)d+2}, \ldots, y_{kd}]^{T}, \no\\
&&\boldsymbol{\eta}_{k, d} = [\eta_{(k-1)d+1}, \eta_{(k-1)d+2}, \ldots, \eta_{kd}]^{T}. \no
\end{eqnarray}

The proposed algorithm adopts batch processing. In particular, the proposed algorithm initializes the first $d$ decisions as a group of standard orthogonal basis. The decision maker updates the estimate $\hat{\boldsymbol{\theta}}_{t}$ whenever he collects $d$ successive rewards. Furthermore, whenever a new estimate $\hat{\boldsymbol{\theta}}_{t}$ is calculated, the decision maker chooses next decision $\mathbf{x}_{t+1}$ as the direction of $\hat{\boldsymbol{\theta}}_{t}$, and selects another $d-1$ decisions such that these $d$ decisions form another group of orthogonal basis. We emphasize that algorithms according to the OFU principle keep taking decisions that maximize the reward $<\mathbf{x}, \hat{\boldsymbol{\theta}}_{t}>$. In our context, the OFU algorithm will always select the decision with the same direction of $\hat{\boldsymbol{\theta}}_{t}$. However, in our proposed algorithm, among every successive $d$ decisions, only one decision is on the direction of $\hat{\boldsymbol{\theta}}_{t}$; the rest of $d-1$ decisions are orthogonal to $\hat{\boldsymbol{\theta}}_{t}$. This is the key difference between the OFU algorithm and our algorithm.

\begin{algorithm}
\KwData{the adaptively designed decisions $\mathbf{x}_{1}, \ldots, \mathbf{x}_{t}$ and corresponding rewards $y_{1}, \ldots, y_{t}$}
\KwResult{the estimate $\hat{\boldsymbol{\theta}}_{t}$}
Initialization: select $\mathbf{x}_{1}, \ldots, \mathbf{x}_{d}$ as a set of standard orthogonal basis \;
\For{ $k = 1, 2, \ldots \lceil t/d \rceil$}{
obtain rewards: $\mathbf{Y}_{k, d} = \mathbf{X}_{k,d} \boldsymbol{\theta^{*}} + \boldsymbol{\eta}_{k, d}$ \;
update matrix: $\mathbf{W}_{kd} = \mathbf{W}_{(k-1)d} +  \mathbf{X}_{k,d}^{T}\mathbf{X}_{k,d}$ \;
estimate parameter: $\hat{\boldsymbol{\theta}}_{kd} = \mathbf{W}_{kd}^{-1}\mathbf{X}_{kd}^{T}\mathbf{Y}_{kd}$ \;
choose decision: $\mathbf{x}_{kd+1} = \hat{\boldsymbol{\theta}}_{kd}/||\hat{\boldsymbol{\theta}}_{kd}||_{2}^{2}$, select $\{\mathbf{x}_{kd+1}, \mathbf{x}_{kd+2}, \ldots, \mathbf{x}_{(k+1)d}\}$ to be an orthogonal basis;
}
\caption{The Proposed Algorithm} \label{alg:proposed}
\end{algorithm}

The performance of the proposed algorithm is characterized in the following theorem.
\begin{thm} \label{thm:proposed_MSE}
For the proposed algorithm, we have
$$\mathbb{E}[|| \hat{\boldsymbol{\theta}}_{t} - \boldsymbol{\theta^{*}} ||_{2}^2] \leq \frac{d^2}{t}\sigma^2 (1+o(1)).$$
Furthermore, if $\boldsymbol{\eta}_{t}$ is a sub-Gaussian vector, then
\begin{eqnarray}
P\left(||\hat{\boldsymbol{\theta}}_{t} - \boldsymbol{\theta^{*}}||_{2}^2 \geq \frac{3\sigma^2 d^{3/2}}{t^{1/2}} + O\left(\frac{\sigma^2 d^2}{t}\right)\right) \leq e^{-t} \no
\end{eqnarray}
\end{thm}
\begin{IEEEproof}
Please see Appendix \ref{adp:proposed performance}.
\end{IEEEproof}

Theorem \ref{thm:proposed_MSE} characterizes our performance metric \eqref{eq:err_prob}. In particular, $\delta$ decays exponentially as $t\rightarrow \infty$, and the bound of estimation error $\epsilon$ shrinks to zero on the order $O(t^{-1/2})$ for the proposed algorithm.
%It is easy to see that $||(\hat{\boldsymbol{\theta}}_{t} - \boldsymbol{\theta})||^2$ decay in order to $t^{-1}$ for normal case by Markov inequality. However, above result shows that it decays exponentially for sub-Gaussian noise.

We now provide a lower bound of the mean square estimation error (MSE) for all possible sequential decision selection strategies and show that MSE reduces at most on order $O(t^{-1})$.
%\begin{eqnarray}
%\hat{\boldsymbol{\theta}}_{t} - \boldsymbol{\theta^{*}} = -\mathbf{W}_{t}^{-1}\mathbf{W}_{0}\boldsymbol{\theta^{*}} + \mathbf{W}_{t}^{-1}\mathbf{X}_{t}^{T}\boldsymbol{\eta}_{t}
%\end{eqnarray}

\begin{thm} \label{thm:lowerbound}
(Lower Bounds on MSE) Let $\eta_{t}$ be a sub-Gaussian random variable with variance proxy $\sigma^2$. % $\mathbf{W}_{0}$ is a $d \times d$ positive definite matrix. Then
If estimator \eqref{eq:estimator} is adopted, then for any adaptively selected decision sequence $\{\mathbf{x}_{i}, i=1, 2, \ldots, t\}$, we have
\begin{eqnarray}
\mathbb{E}[|| \hat{\boldsymbol{\theta}}_{t} - \boldsymbol{\theta^{*}} ||_{2}^2] \geq \frac{1}{t} \sigma^2 + o\left(\frac{1}{t}\right).
\end{eqnarray}
\end{thm}
\begin{IEEEproof}
Please see Appendix \ref{adp:lowerbound}.
\end{IEEEproof}

Theorem \ref{thm:proposed_MSE} indicates that the convergence rate of MSE for the proposed algorithm is on order $O(t^{-1})$, while Theorem~\ref{thm:lowerbound} shows that the convergence rate of MSE cannot be faster than $O(t^{-1})$. Hence, the proposed algorithm is order optimal.

\section{Numerical Simulation} \label{sec:sim}
\begin{figure}[h]
\begin{center}
\includegraphics[width=8cm]{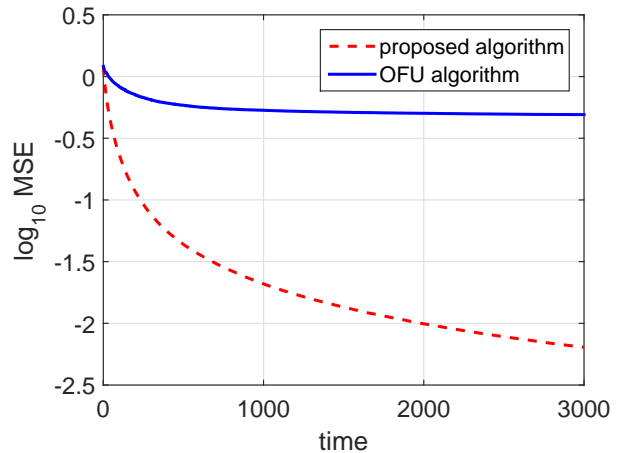}
\caption{Estimation error vs. the total number of decisions}
\label{fig:ADD_vs_Pfa_1}
\end{center}
\end{figure}

In this section, we provide a numerical example to illustrate the results obtained in this paper. In this numerical example, we set $d=5$, %and we randomly create an underlying parameter $\boldsymbol{\theta}^{*}$. In the simulation,
and compare the performance of the OFU algorithm and our proposed algorithm. In particular, the MSE of each algorithm is calculated by Monte Carlo method. In the simulation, the estimation procedure proceeds $3000$ rounds; hence, for each trial, the decision maker has to adaptively make $3000$ decisions. For each algorithm, we conduct $10^5$ trials with randomly created underlying parameter $\boldsymbol{\theta}^{*}$, and we record the estimation error at each round of decision. Then, the logarithm of MSE, which is estimated by the average of estimation error at each trial, at each decision round is illustrated in Figure \ref{fig:ADD_vs_Pfa_1}.

%Figure \ref{fig:ADD_vs_Pfa_1} reflects the relationship between the logarithm of the mean square estimation error and the number of decisions.
In Figure \ref{fig:ADD_vs_Pfa_1}, The blue solid line is the performance of the OFU algorithm and the red dash line is the performance of the proposed algorithm. The simulation result shows that the error of the OFU algorithm tends to be a constant as the number of decisions goes large; hence, the corresponding MSE also tends to a constant. However, the error of the proposed algorithm decays when the number of decisions grows, which indicates the estimation error tends to zero as the number of decisions goes to infinity. Hence, the proposed estimator is consistent. %Theorem \ref{thm:proposed_MSE} also indicates that the red curve has the shape of $-\log t$.

\section{Conclusion} \label{sec:con}
In this paper, we have studied the problem of identifying the best action in the stochastic linear bandit setup with a fixed confidence constraint. We have shown that the existing OFU algorithm is an inconsistent estimator for the unknown parameter $\boldsymbol{\theta}^{*}$. We have proposed and analyzed a novel algorithm. We have shown that the proposed algorithm is consistent and that its mean square estimation error reduces on order $O(t^{-1})$. Furthermore, we have shown that the probability that the estimation error is larger than $t^{-1/2}$ decays exponentially with respect to $t$.

We note that this paper has considered the asymptotic case with $t\rightarrow\infty$. In the future, it will be of interest to consider the problem when a finite number of decisions are made. In this case, we expect that tools from optimal stopping~\cite{Poor:Book:08} and controlled sensing~\cite{Nitinawarat:TAC:13} will be useful.

\bibliographystyle{ieeetr}{}
\bibliography{macros,detection,bandit}

\begin{thebibliography}{10}

\bibitem{Press:NAS:09}
W.~H. Press, ``Bandit solutions provide unified ethical models for randomized
  clinical trials and comparative effecctiveness research,'' {\em Proceedings
  of the National Academy of Sciences}, vol.~106, pp.~22387--22392, Dec. 2009.

\bibitem{Awerbuch:STOC:04}
B.~Awerbuch and P.~Kleinberg, ``Adaptive routing with end-to-end feedback:
  Distributed learning and geometric approaches,'' in {\em Proc. Annual ACM
  Symp. Theory of Computing}, (Chicago, IL), pp.~45--53, June 2004.

\bibitem{Manickam:ICASSP:17}
I.~Manickam, A.~S. Lan, and R.~G. Baraniuk, ``Contextual multi-armed bandit
  algorithms for personalized learning action selection,'' in {\em Proc. IEEE
  Intl. Conf. on Acoustics, Speech, and Signal Processing}, (New Orleans, LA),
  pp.~6344--6348, June 2017.

\bibitem{Reverdy:ICASSP:18}
P.~Reverdy and V.~Srivastava, ``Multi-armed bandits for human-machine decision
  making,'' in {\em Proc. IEEE Intl. Conf. on Acoustics, Speech, and Signal
  Processing}, (Calagry, CA), pp.~6986--6990, Apr. 2018.

\bibitem{Lai:TMC:11}
L.~Lai, H.~E. Gamal, H.~Jiang, and H.~V. Poor, ``Cognitive medium access:
  Exploration, exploitation, and competition,'' {\em IEEE Transactions on
  Mobile Computing}, vol.~10, pp.~239--253, Feb 2011.

\bibitem{Gabillon:NIPS:12}
V.~Gabillon, M.~Ghavamzadeh, and A.~Lazaric, ``Best arm identification: A
  unified approach to fixed budget and fixed confidence,'' in {\em Advances in
  Neural Information Processing Systems}, (Lake Tahoe, USA), pp.~1--16, Dec.
  2012.

\bibitem{Jamieson:CISS:14}
K.~Jamieson and R.~Nowak, ``Best-arm indentification algorithms for multi-armed
  bandits in the fixed confidence setting,'' in {\em Proc. Conf. on Information
  Science and Systems}, (Princeton, NJ), pp.~1--6, Mar. 2014.

\bibitem{Jamieson:JMLR:14}
K.~Jamieson, M.~Malloy, R.~Nowak, and S.~Bubeck, ``lil'ucb: An optimal
  exploration algorithm for multi-armed bandits,'' {\em Journal of Machine
  Learning Research}, vol.~35, pp.~1--17, Dec. 2014.

\bibitem{Karnin:ICML:13}
Z.~Karnin, T.~Koren, and S.~Somekh, ``Almost optimal exploration in multi-armed
  bandits,'' in {\em Proc. Intl. Conf. on Machine Learning}, June 2013.

\bibitem{Jamieson:arXiv:13}
K.~Jamieson, M.~Malloy, R.~Nowak, and S.~Bubeck, ``On finding the largest mean
  among many,'' {\em arXiv preprint arXiv:1306.3917}, 2013.

\bibitem{Auer:JMLR:02}
P.~Auer, ``Using confidence bounds for exploitation-exploration trade-offs,''
  {\em Journal of Machine Learning Research}, vol.~3, pp.~397--422, 2002.

\bibitem{Kalyanakrishnan:ICML:12}
S.~Kalyanakrishnan, A.~Tewari, and P.~Stone, ``{PAC} subset selection in
  stochastic multi-armed bandits,'' in {\em Proc. Intl. Conf. on Machine
  Learning}, pp.~655--662, June 2012.

\bibitem{Dani:COLT:08}
V.~Dani, T.~P. Hayes, and S.~M. Kakade, ``Stochastic linear optimization under
  bandit feedback,'' in {\em Proc. Annual Conference on Learning Theory},
  pp.~355--366, 2008.

\bibitem{Abbasi:NIPS:11}
Y.~Abbasi-Yadkori, D.~Pal, and C.~Szepesvari, ``Improved algorithms for linear
  stochastic bandits,'' in {\em Advances in Neural Information Processing
  Systems}, (Granada, Spain), pp.~1--19, Dec. 2011.

\bibitem{Poor:Book:08}
H.~V. Poor and O.~Hadjiliadis, {\em Quickest Detection}.
\newblock Cambridge, UK: Cambridge University Press, 2008.

\bibitem{Nitinawarat:TAC:13}
S.~Nitinawarat, G.~K. Atia, and V.~V. Veeravalli, ``Controlled sensing for
  multihypothesis testing,'' {\em IEEE Trans. Automatic Control}, vol.~58,
  pp.~2451-- 2464, Oct. 2013.

\bibitem{Hsu:ECP:12}
D.~Hsu, S.~M. Kakade, and T.~Zhang, ``A tail inequality for quadratic forms of
  subgaussian random vectors,'' {\em Electronic Communications in Probability},
  vol.~17, pp.~1--6, 2012.

\end{thebibliography}
\appendices

\newpage
\section{Proof of the Theorem \ref{thm:OFU_error}} \label{adp:OFU_error}
Recall that the decision set is
$$ D_{t} =\{\mathbf{x} \in \mathbb{R}^{d}: ||\mathbf{x}||_{2}^{2} \leq 1\}, $$
hence, for the OFU algorithm, it is easy to see that the decision selected by the decision maker is
\begin{eqnarray}
\mathbf{x}_{t+1} = \text{argmax}_{\mathbf{x}\in D_{t}} <\mathbf{x}, \hat{\boldsymbol{\theta}}_{t}> = \hat{\boldsymbol{\theta}}_{t}/||\hat{\boldsymbol{\theta}}_{t}||_{2}^{2}, \no
\end{eqnarray}
and the optimal decision with known $\thetav^*$ is
\begin{eqnarray}
\mathbf{x}^{*} = \text{argmax}_{\mathbf{x}\in D_{t}} <\mathbf{x}, \boldsymbol{\theta}^{*}> = \boldsymbol{\theta}^{*}/||\boldsymbol{\theta}^{*}||_{2}^{2}. \no
\end{eqnarray}
For notation convenience, we denote $\underline{\hat{\boldsymbol{\theta}}}_{t} := \hat{\boldsymbol{\theta}}_{t}/||\hat{\boldsymbol{\theta}}_{t}||_{2}^{2}$ and $\underline{\boldsymbol{\theta}} = \boldsymbol{\theta}^{*}/||\boldsymbol{\theta}^{*}||_{2}^2$. Then the cumulative regret can be written as
\begin{eqnarray}
R_{n} = \sum_{t=1}^{n} <\mathbf{x}_{t}^{*} - \mathbf{x_{t}}, \boldsymbol{\theta}^{*}>  = ||\boldsymbol{\theta}^{*}||_{2}^{2}\sum_{t=1}^{n} <\underline{\boldsymbol{\theta}} - \underline{\boldsymbol{\hat{\theta}}}_{t-1}, \underline{\boldsymbol{\theta}}>, \no
\end{eqnarray}
and the assumption $\lim_{n\rightarrow \infty} R_{n}/n = 0$ indicates
\begin{eqnarray}
\lim_{n\rightarrow \infty} \frac{1}{n}\sum_{t=1}^{n} <\underline{\boldsymbol{\theta}} - \underline{\boldsymbol{\hat{\theta}}}_{t-1}, \underline{\boldsymbol{\theta}}> = 0. \label{eq:hannan}
\end{eqnarray}
In the following, we calculate the lower bound of the estimation error. Since
\begin{eqnarray}
\hat{\boldsymbol{\theta}}_{t} = \mathbf{W}_{t}^{-1}\mathbf{X}_{t}^{T}\mathbf{Y}_{t}, \no
\end{eqnarray}
we have
\begin{eqnarray}
\hat{\boldsymbol{\theta}}_{t} - \boldsymbol{\theta^{*}} &=& \mathbf{W}_{t}^{-1}\mathbf{X}_{t}^{T}(\mathbf{X}_{t}^{T}\boldsymbol{\theta^{*}}+\boldsymbol{\eta}_{t}) - \boldsymbol{\theta^{*}} \no\\
&=& -\mathbf{W}_{t}^{-1}\mathbf{W}_{0}\boldsymbol{\theta^{*}} + \mathbf{W}_{t}^{-1}\mathbf{X}_{t}^{T}\boldsymbol{\eta}_{t},
\end{eqnarray}
in which we have used~\eqref{eq:WtW0}.

Therefore,
\begin{eqnarray}
\mathbf{W}_{t}^{1/2}\left(\hat{\boldsymbol{\theta}}_{t} - \boldsymbol{\theta^{*}}\right) = \mathbf{W}_{t}^{-1/2}\left(\mathbf{X}_{t}^{T}\boldsymbol{\eta}_{t} - \mathbf{W}_{0}\boldsymbol{\theta^{*}}\right), \no
\end{eqnarray}
and we have
\begin{eqnarray}
(\hat{\boldsymbol{\theta}}_{t} - \boldsymbol{\theta^{*}})^{T}\mathbf{W}_{t}(\hat{\boldsymbol{\theta}}_{t} - \boldsymbol{\theta^{*}}) = ||\mathbf{X}_{t}^{T}\boldsymbol{\eta}_{t} - \mathbf{W}_{0}\boldsymbol{\theta^{*}}||_{\mathbf{W}_{t}^{-1}}^{2}. \no
\end{eqnarray}

As a result, we have
\begin{eqnarray}
&&\left[(\hat{\boldsymbol{\theta}}_{t} - \boldsymbol{\theta^{*}})^{T}\mathbf{W}_{t}(\hat{\boldsymbol{\theta}}_{t} - \boldsymbol{\theta^{*}})\right]^{1/2} \no\\
&&\geq ||\mathbf{X}_{t}^{T}\boldsymbol{\eta}_{t}||_{\mathbf{W}_{t}^{-1}} - ||\mathbf{W}_{0}\boldsymbol{\theta^{*}}||_{\mathbf{W}_{t}^{-1}} \no\\
&&\geq ||\mathbf{X}_{t}^{T}\boldsymbol{\eta}_{t}||_{\mathbf{W}_{t}^{-1}} - \frac{\lambda_{\max}(\mathbf{W}_{0})}{\lambda_{\min}^{1/2}(\mathbf{W}_{0})}||\boldsymbol{\theta^{*}}||_{2}, \label{eq:lb1}
\end{eqnarray}
%hence, for any vector $\mathbf{v}$
%\begin{eqnarray}
%&&|\mathbf{v}^{T}(\hat{\boldsymbol{\theta}}_{t} - \boldsymbol{\theta^{*}})| \no\\
%&&= |<\mathbf{v}, \mathbf{X}_{t}^{T}\boldsymbol{\eta}_{t}>_{\mathbf{W}_{t}^{-1}} - <\mathbf{v}, \mathbf{W}_{0}\boldsymbol{\theta^{*}}>_{\mathbf{W}_{t}^{-1}}| \no\\
%&&\geq ||\mathbf{v}||_{\mathbf{W}_{t}^{-1}} \left( ||\mathbf{X}_{t}^{T}\boldsymbol{\eta}_{t}||_{\mathbf{W}_{t}^{-1}} - ||\mathbf{W}_{0}\boldsymbol{\theta^{*}}||_{\mathbf{W}_{t}^{-1}} \right) \no\\
%&&\geq ||\mathbf{v}||_{\mathbf{W}_{t}^{-1}} \left( ||\mathbf{X}_{t}^{T}\boldsymbol{\eta}_{t}||_{\mathbf{W}_{t}^{-1}} - \frac{\lambda_{\max}(\mathbf{W}_{0})^{2}}{\lambda_{\min}(\mathbf{W}_{0})}||\boldsymbol{\theta^{*}}||_{2}^{2} \right), \no
%\end{eqnarray}
in which the last step is true since $$||\mathbf{W}_{0}\boldsymbol{\theta}^{*}||_{\mathbf{W}_{t}^{-1}}^{2} \leq \frac{1}{\lambda_{\min}{(\mathbf{W}_{t})}}||\mathbf{W}_{0}\boldsymbol{\theta^{*}}||_{2}^{2} \leq \frac{\lambda_{\max}(\mathbf{W}_{0}^{2})}{\lambda_{\min}(\mathbf{W}_{0})}||\boldsymbol{\theta^{*}}||_{2}^{2}. $$
%By setting $\mathbf{v} = \mathbf{W}_{t}(\hat{\boldsymbol{\theta}}_{t} - \boldsymbol{\theta^{*}})$, we then have
%\begin{eqnarray}
%&&\left[(\hat{\boldsymbol{\theta}}_{t} - \boldsymbol{\theta^{*}})^{T}\mathbf{W}_{t}(\hat{\boldsymbol{\theta}}_{t} - \boldsymbol{\theta^{*}})\right]^{1/2} \no\\
%&&\hspace{6mm}\geq ||\mathbf{X}_{t}^{T}\boldsymbol{\eta}_{t}||_{\mathbf{W}_{t}^{-1}} - \frac{\lambda_{\max}(\mathbf{W}_{0})^{2}}{\lambda_{\min}(\mathbf{W}_{0})}||\boldsymbol{\theta^{*}}||_{2}^{2}. \label{eq:lb1}
%\end{eqnarray}
We note that
\begin{eqnarray}
&&\mathbf{W}_{t} - t\underline{\boldsymbol{\theta}}\underline{\boldsymbol{\theta}}^{T} \no\\
&&= \mathbf{W}_{0} + \sum_{i=1}^{t} \mathbf{x}_{i}\mathbf{x}_{i}^{T} - t\underline{\boldsymbol{\theta}}\underline{\boldsymbol{\theta}}^{T} \no\\
&&= \mathbf{W}_{0} + (\mathbf{x}_{1}\mathbf{x}_{1}^{T} - \underline{\boldsymbol{\theta}}\underline{\boldsymbol{\theta}}^{T}) + \sum_{i=1}^{t-1} \left(\underline{\hat{\boldsymbol{\theta}}}_{i}\underline{\hat{\boldsymbol{\theta}}}_{i}^{T} - \underline{\boldsymbol{\theta}}\underline{\boldsymbol{\theta}}^{T}\right) \no\\
&&= \mathbf{W}_{0} + (\mathbf{x}_{1}\mathbf{x}_{1}^{T} - \underline{\boldsymbol{\theta}}\underline{\boldsymbol{\theta}}^{T}) + \sum_{i=1}^{t-1} \left(\underline{\hat{\boldsymbol{\theta}}}_{i} + \underline{\boldsymbol{\theta}} \right)\left(\underline{\hat{\boldsymbol{\theta}}}_{i} - \underline{\boldsymbol{\theta}}\right)^{T} \no\\
&&= \mathbf{W}_{0} + (\mathbf{x}_{1}\mathbf{x}_{1}^{T} - \underline{\boldsymbol{\theta}}\underline{\boldsymbol{\theta}}^{T}) + \sum_{i=1}^{t-1} 2\underline{\boldsymbol{\theta}}\left(\underline{\hat{\boldsymbol{\theta}}}_{i} - \underline{\boldsymbol{\theta}}\right)^{T} \no\\
&&+ \sum_{i=1}^{t-1}  \left(\underline{\hat{\boldsymbol{\theta}}}_{i} - \underline{\boldsymbol{\theta}} \right)\left(\underline{\hat{\boldsymbol{\theta}}}_{i} - \underline{\boldsymbol{\theta}}\right)^{T}. \no
\end{eqnarray}
Then, we have
\begin{eqnarray}
\mathbf{W}_{t} &=& \mathbf{W}_{0} + (\mathbf{x}_{1}\mathbf{x}_{1}^{T} - \underline{\boldsymbol{\theta}}\underline{\boldsymbol{\theta}}^{T}) + t\underline{\boldsymbol{\theta}}\underline{\boldsymbol{\theta}}^{T} \no\\
&+& \sum_{i=1}^{t-1} 2\underline{\boldsymbol{\theta}}\left(\underline{\hat{\boldsymbol{\theta}}}_{i} - \underline{\boldsymbol{\theta}}\right)^{T} + \sum_{i=1}^{t-1}  \left(\underline{\hat{\boldsymbol{\theta}}}_{i} - \underline{\boldsymbol{\theta}} \right)\left(\underline{\hat{\boldsymbol{\theta}}}_{i} - \underline{\boldsymbol{\theta}}\right)^{T}. \no
\end{eqnarray}
Therefore,
\begin{eqnarray}
\frac{1}{t-1}\mathbf{W}_{t}\underline{\boldsymbol{\theta}} &=& \frac{1}{t-1}(\mathbf{W}_{0}+\mathbf{x}_{1}\mathbf{x}_{1}^{T} )\underline{\boldsymbol{\theta}} + \underline{\boldsymbol{\theta}}\underline{\boldsymbol{\theta}}^{T}\underline{\boldsymbol{\theta}} \no\\
&+& \frac{1}{t-1}\sum_{i=1}^{t-1} 2\underline{\boldsymbol{\theta}}\left(\underline{\hat{\boldsymbol{\theta}}}_{i}\underline{\boldsymbol{\theta}} - \underline{\boldsymbol{\theta}}\right)^{T} \no\\
&+& \frac{1}{t-1}\sum_{i=1}^{t-1}  \left(\underline{\hat{\boldsymbol{\theta}}}_{i} - \underline{\boldsymbol{\theta}} \right)\left(\underline{\hat{\boldsymbol{\theta}}}_{i} - \underline{\boldsymbol{\theta}}\right)^{T}\underline{\boldsymbol{\theta}}. \no
\end{eqnarray}
As $t\rightarrow \infty$, the first item on the right hand side of the equality
\begin{eqnarray}
&&\frac{1}{t-1}(\mathbf{W}_{0}+\mathbf{x}_{1}\mathbf{x}_{1}^{T})\underline{\boldsymbol{\theta}} \rightarrow \mathbf{0} \no
\end{eqnarray}
because $\mathbf{x}_{1}$ and $\underline{\boldsymbol{\theta}}$ have finite norms. The third item
\begin{eqnarray}
&&\frac{1}{t-1} \sum_{i=1}^{t-1} 2\underline{\boldsymbol{\theta}}\left(\underline{\hat{\boldsymbol{\theta}}}_{i} - \underline{\boldsymbol{\theta}}\right)^{T}\underline{\boldsymbol{\theta}} \no\\
&&\hspace{10mm}= 2\underline{\boldsymbol{\theta}}\left[ \frac{1}{t-1} \sum_{i=1}^{t-1} <\hat{\underline{\boldsymbol{\theta}}}_{i} - \underline{\boldsymbol{\theta}},  \underline{\boldsymbol{\theta}}> \right] {\rightarrow} \mathbf{0} \no
\end{eqnarray}
and the forth item
\begin{eqnarray}
&&\frac{1}{t-1}\sum_{i=1}^{t-1}  \left(\underline{\hat{\boldsymbol{\theta}}}_{i} - \underline{\boldsymbol{\theta}} \right)\left(\underline{\hat{\boldsymbol{\theta}}}_{i} - \underline{\boldsymbol{\theta}}\right)^{T}\underline{\boldsymbol{\theta}} \no\\
&&= \frac{1}{t-1}\sum_{i=1}^{t-1} (\hat{\underline{\boldsymbol{\theta}}}_{i} - \underline{\boldsymbol{\theta}}) <\hat{\underline{\boldsymbol{\theta}}}_{i} - \underline{\boldsymbol{\theta}},  \underline{\boldsymbol{\theta}}> \no\\
&&\leq \max_{i\in\{1, \ldots, t-1\}} ||\hat{\underline{\boldsymbol{\theta}}}_{i} - \underline{\boldsymbol{\theta}}||\frac{1}{t-1} \sum_{i=1}^{t-1} <\hat{\underline{\boldsymbol{\theta}}}_{i} - \underline{\boldsymbol{\theta}},  \underline{\boldsymbol{\theta}}> \no\\
&&\leq \frac{2}{t-1} \sum_{i=1}^{t-1} <\hat{\underline{\boldsymbol{\theta}}}_{i} - \underline{\boldsymbol{\theta}},  \underline{\boldsymbol{\theta}}> {\rightarrow} \mathbf{0}
\end{eqnarray}
because of \eqref{eq:hannan}. Since \eqref{eq:hannan} holds with probability at least $1-\delta$, then as $t \rightarrow \infty$,
$$ \frac{1}{t-1}\mathbf{W}_{t}\underline{\boldsymbol{\theta}} = \underline{\boldsymbol{\theta}} $$
holds with probability $1-\delta$. That is, $\underline{\boldsymbol{\theta}}$ is the eigenvector associated with eigenvalue 1 for matrix $\frac{1}{t-1}\mathbf{W}_{t}$. As $t\rightarrow\infty$, we further have, with probability at least $1-\delta$,
\begin{eqnarray}
%(\hat{\boldsymbol{\theta}}_{t} - \boldsymbol{\theta})^{T} \frac{1}{t-1}\mathbf{W}_{t} (\hat{\boldsymbol{\theta}}_{t} - \underline{\boldsymbol{\theta}}) &=&
&&\frac{1}{t-1} \boldsymbol{\eta}_{t}^{T} \mathbf{X}_{t} \mathbf{W}_{t}^{-1} \mathbf{X}_{t}^{T} \boldsymbol{\eta}_{t} \no\\
&&= \frac{1}{(t-1)^2} \boldsymbol{\eta}_{t}^{T} \mathbf{X}_{t} \left(\frac{1}{t-1}\mathbf{W}_{t}\right)^{-1} \mathbf{X}_{t}^{T} \boldsymbol{\eta}_{t} \no\\
&&\overset{(a)}{\geq} \frac{1}{(t-1)^2} \boldsymbol{\eta}_{t}^{T} \mathbf{X}_{t} \left(\underline{\boldsymbol{\theta}}\underline{\boldsymbol{\theta}}^{T}\right) \mathbf{X}_{t}^{T} \boldsymbol{\eta}_{t} \no\\
&&= \frac{1}{(t-1)^2} ( \boldsymbol{\eta}_{t}^{T} \mathbf{X}_{t} \underline{\boldsymbol{\theta}})^2 \no\\
&&= \frac{1}{(t-1)^2} ( \text{trace}( \mathbf{X}_{t} \underline{\boldsymbol{\theta}}\boldsymbol{\eta}_{t}^{T}))^2 \no\\
&&\overset{(b)}{=} \frac{1}{(t-1)^2} \text{trace}( \mathbf{X}_{t} \underline{\boldsymbol{\theta}}\boldsymbol{\eta}_{t}^{T}\boldsymbol{\eta}_{t} \underline{\boldsymbol{\theta}}^{T}\mathbf{X}_{t}^{T} )\no\\
&&= \frac{1}{t-1} \text{trace}\left( \mathbf{X}_{t} \underline{\boldsymbol{\theta}} \left(\frac{1}{t-1}\boldsymbol{\eta}_{t}^{T}\boldsymbol{\eta}_{t}\right) \underline{\boldsymbol{\theta}}^{T}\mathbf{X}_{t}^{T} \right)\no\\
&&\overset{(c)}{=} \frac{\sigma^2}{t-1} \text{trace}( \mathbf{X}_{t} \underline{\boldsymbol{\theta}} \underline{\boldsymbol{\theta}}^{T}\mathbf{X}_{t}^{T} )\no\\
&&= \frac{\sigma^2}{t-1} \text{trace}( \underline{\boldsymbol{\theta}}^{T}\mathbf{X}_{t}^{T} \mathbf{X}_{t} \underline{\boldsymbol{\theta}} )\no\\
&&\overset{(d)}{=} \sigma^2. \label{eq:lb2}
\end{eqnarray}
In above derivations, $\frac{1}{t-1}\mathbf{W}_{t}$ is a positive definite matrix, then $(\frac{1}{t-1}\mathbf{W}_{t})^{-1}$ is a positive definite matrix sharing the same eigenvectors with $\frac{1}{t-1}\mathbf{W}_{t}$. Hence (a) holds because $(\frac{1}{t-1}\mathbf{W}_{t})^{-1}-\underline{\boldsymbol{\theta}}\underline{\boldsymbol{\theta}}^{T} \succeq \mathbf{0}$. (b) is true, becuase for a rank 1 matrix $\mathbf{A}$, we have $\text{trace}[\mathbf{A}\mathbf{A}^{T}] = \lambda(\mathbf{A})^2 = \text{trace}[\mathbf{A}]^2$. (c) is true because $\lim_{t\rightarrow\infty}\frac{1}{t-1} \boldsymbol{\eta}_{t}^{T}\boldsymbol{\eta}_{t} = \frac{1}{t-1} \sum_{i=1}^{t}\eta_{i}^{2} = \sigma^2$ holds almost surely under the strong law of large number. (d) is true because
\begin{eqnarray}
\underline{\boldsymbol{\theta}}^{T}\frac{1}{t-1}\mathbf{X}_{t}^{T} \mathbf{X}_{t} \underline{\boldsymbol{\theta}} &=& \underline{\boldsymbol{\theta}}^{T}\frac{1}{t-1}(\mathbf{W}_{t} - \mathbf{W}_{0})\underline{\boldsymbol{\theta}} \no\\
&=& \underline{\boldsymbol{\theta}}^{T}\frac{1}{t-1}\mathbf{W}_{t}\underline{\boldsymbol{\theta}} - \underline{\boldsymbol{\theta}}^{T}\frac{1}{t-1}\mathbf{W}_{0}\underline{\boldsymbol{\theta}} \no\\
&=& \underline{\boldsymbol{\theta}}^{T} \underline{\boldsymbol{\theta}} - \frac{1}{t-1}\underline{\boldsymbol{\theta}}^{T}\mathbf{W}_{0}\underline{\boldsymbol{\theta}} = 1. \no
\end{eqnarray}
We also have
\begin{eqnarray}
&&(\hat{\boldsymbol{\theta}}_{t} - \boldsymbol{\theta^{*}})^{T} \frac{1}{t-1}\mathbf{W}_{t} (\hat{\boldsymbol{\theta}}_{t} - \boldsymbol{\theta^{*}}) \no\\
&&\leq \lambda_{\max}(\frac{1}{t-1}\mathbf{W}_{t}) (\hat{\boldsymbol{\theta}}_{t} - \boldsymbol{\theta^{*}})^{T}  (\hat{\boldsymbol{\theta}}_{t} - \boldsymbol{\theta^{*}}) \no\\
&&\leq \text{trace}(\frac{1}{t-1}\mathbf{W}_{t}) (\hat{\boldsymbol{\theta}}_{t} - \boldsymbol{\theta^{*}})^{T}(\hat{\boldsymbol{\theta}}_{t} - \boldsymbol{\theta^{*}}) \leq ||\hat{\boldsymbol{\theta}}_{t} - \boldsymbol{\theta^{*}}||^2. \no
\end{eqnarray}
Therefore we have
\begin{eqnarray}
&&||\hat{\boldsymbol{\theta}}_{t} - \boldsymbol{\theta^{*}}||_{2}^2 \no\\
&&\geq (\hat{\boldsymbol{\theta}}_{t} - \boldsymbol{\theta^{*}})^{T} \frac{1}{t-1}\mathbf{W}_{t} (\hat{\boldsymbol{\theta}}_{t} - \boldsymbol{\theta^{*}}) \no\\
&&\overset{(a)}\geq \frac{1}{t-1}\left[ ||\mathbf{X}_{t}^{T}\boldsymbol{\eta}_{t}||_{\mathbf{W}_{t}^{-1}} - \frac{\lambda_{\max}(\mathbf{W}_{0})}{\lambda_{\min}^{1/2}(\mathbf{W}_{0})}||\boldsymbol{\theta^{*}}||_{2} \right]^2 \no\\
&&\overset{(b)}{=} \frac{1}{t-1} \boldsymbol{\eta}_{t}^{T}\mathbf{X}_{t}\mathbf{W}_{t}^{-1}\mathbf{X}_{t}^{T}\boldsymbol{\eta}_{t} = \sigma^2 \label{eq:lb}
\end{eqnarray}
with probability at least $1-\delta$. In \eqref{eq:lb}, (a) and (b) are due to \eqref{eq:lb1} and \eqref{eq:lb2} respectively.

\section{Proof of Theorem \ref{thm:proposed_MSE} } \label{adp:proposed performance}
In this appendix, we show Theorem \ref{thm:proposed_MSE}. %and Theorem \ref{thm:proposed_prob}. %we first show Theorem \ref{thm:proposed_MSE}.
Recall that
\begin{eqnarray}
%&&\mathbf{X}_{k,d} = \left( \begin{array}{c} \mathbf{x}_{d(k-1)+1}^{T} \\ \mathbf{x}_{d(k-1)+2}^{T} \\ \cdots \\ \mathbf{x}_{dk}^{T} \end{array}\right),  \no\\
&&\mathbf{X}_{k,d} = \left[ \mathbf{x}_{(k-1)d+1}^{T}, \mathbf{x}_{(k-1)d+2}^{T}, \cdots, \mathbf{x}_{kd}^{T} \right]^{T}, \no\\
&&\mathbf{Y}_{k, d} = [y_{d(k-1)+1}, y_{d(k-1)+2}, \ldots, y_{dk}]^{T}, \no\\
&&\boldsymbol{\eta}_{k, d} = [\eta_{d(k-1)+1}, \eta_{d(k-1)+2}, \ldots, \eta_{dk}]^{T}. \no
\end{eqnarray}
Since $\mathbf{x}_{(k-1)d+1}, \ldots, \mathbf{x}_{kd}$ are selected as orthogonal basis, then we have $\mathbf{X}_{k,d}\mathbf{X}_{k,d}^T = \mathbf{I}$. Furthermore, it is easy to see that $\mathbf{X}_{k,d}$ is independent of $\boldsymbol{\eta}_{k, d}$.

Let $t=ld$, then
%\begin{eqnarray}
%\hat{\boldsymbol{\theta}}_{t} - \boldsymbol{\theta} = \left(\mathbf{X}_{t}^{T}\mathbf{X}_{t} \right)^{-1}\mathbf{X}_{t}^{T}\boldsymbol{\eta}_{t} = \left(\sum_{k=1}^{l} \mathbf{X}_{k,d}\mathbf{X}_{k,d}^T \right)^{-1} \mathbf{X}_{t}^{T}\boldsymbol{\eta}_{t} = \frac{1}{l} \mathbf{X}_{t}^{T}\boldsymbol{\eta}_{t}
%\end{eqnarray}
\begin{eqnarray}
\mathbf{W}_{t} &=& \mathbf{W}_{0} + \mathbf{X}_{t}^{T}\mathbf{X}_{t} \no\\
&=& \mathbf{W}_{0} + \sum_{k=1}^{l} \mathbf{X}_{k,d}\mathbf{X}_{k,d}^T = \mathbf{W}_{0} + l \mathbf{I}. \no
\end{eqnarray}
Since
$\hat{\boldsymbol{\theta}}_{t} - \boldsymbol{\theta^{*}} = -\mathbf{W}_{t}^{-1}\mathbf{W}_{0}\boldsymbol{\theta^{*}} + \mathbf{W}_{t}^{-1}\mathbf{X}_{t}^{T}\boldsymbol{\eta}_{t}$,
we have
\begin{eqnarray}
&&||\hat{\boldsymbol{\theta}}_{t} - \boldsymbol{\theta^{*}}||_{2}^{2} \no\\
&&= \left(-\mathbf{W}_{t}^{-1}\mathbf{W}_{0}\boldsymbol{\theta^{*}} + \mathbf{W}_{t}^{-1}\mathbf{X}_{t}^{T}\boldsymbol{\eta}_{t}\right)^{T} \no\\
&&\hspace{4mm}\left(-\mathbf{W}_{t}^{-1}\mathbf{W}_{0}\boldsymbol{\theta^{*}} + \mathbf{W}_{t}^{-1}\mathbf{X}_{t}^{T}\boldsymbol{\eta}_{t}\right) \no\\
&&= \boldsymbol{\theta^{*}}^{T}\mathbf{W}_{0}\mathbf{W}_{t}^{-2}\mathbf{W}_{0}\boldsymbol{\theta^{*}} + \boldsymbol{\eta}_{t}^{T}\mathbf{X}_{t}\mathbf{W}_{t}^{-2}\mathbf{X}_{t}^{T}\boldsymbol{\eta}_{t} \no\\
&&\hspace{4mm}- 2\boldsymbol{\theta^{*}}^{T}\mathbf{W}_{0}\mathbf{W}_{t}^{-2}\mathbf{X}_{t}^{T}\boldsymbol{\eta}_{t} \no\\
&&\overset{(a)}\leq \frac{1}{l^2}\boldsymbol{\theta^{*}}^{T}\mathbf{W}_{0}\mathbf{W}_{0}\boldsymbol{\theta^{*}} + \frac{1}{l^2}\boldsymbol{\eta}_{t}^{T}\mathbf{X}_{t}\mathbf{X}_{t}^{T}\boldsymbol{\eta}_{t} \no\\
&&\hspace{4mm}- 2\boldsymbol{\theta^{*}}^{T}\mathbf{W}_{0}\mathbf{W}_{t}^{-2}\mathbf{X}_{t}^{T}\boldsymbol{\eta}_{t}, \label{eq:upperBound}
\end{eqnarray}
in which (a) is because of $\mathbf{W}_{t}^{-2} \preceq l^{-2} \mathbf{I}$. To see this, note that $\mathbf{W}_{t} - l\mathbf{I} = \mathbf{W}_{0} \succeq \mathbf{0}$, hence we can obtain $\mathbf{W}_{t} \succeq l\mathbf{I}$, which further indicates $\mathbf{W}_{t}^{-1} \preceq l^{-1} \mathbf{I}. $

We then calculate the expectations of the three items in the right hand side of \eqref{eq:upperBound}. We note that the first item $\boldsymbol{\theta^{*}}^{T}\mathbf{W}_{0}\mathbf{W}_{0}\boldsymbol{\theta^{*}}$ is a constant. For the second item, we have
\begin{eqnarray}
&&\frac{1}{l^{2}}\mathbb{E}[\boldsymbol{\eta}_{t}^{T}\mathbf{X}_{t}\mathbf{X}_{t}^{T}\boldsymbol{\eta}_{t}] \no\\
&&= \frac{1}{l^{2}}\mathbb{E}\left[\sum_{i=1}^{l}\boldsymbol{\eta}_{i, d}^{T}\mathbf{X}_{i, d} \sum_{j=1}^{l}\mathbf{X}_{j, d}^{T}\boldsymbol{\eta}_{j, d}\right] \no\\
&&= \frac{1}{l^{2}} \sum_{i=1}^{l}\sum_{j=1}^{l} \mathbb{E}\left[\boldsymbol{\eta}_{i, d}^{T}\mathbf{X}_{i, d} \mathbf{X}_{j, d}^{T}\boldsymbol{\eta}_{j, d}\right] \no\\
&&= \frac{1}{l^{2}} \sum_{i=1}^{l}\sum_{j=1}^{l} \mathbb{E}\left[\text{trace}[\mathbf{X}_{j, d}^{T}\boldsymbol{\eta}_{j, d}\boldsymbol{\eta}_{i, d}^{T}\mathbf{X}_{i, d}]\right] \no\\
&&= \frac{1}{l^{2}} \sum_{i=1}^{l}\sum_{j=1}^{l} \text{trace}\left[\mathbb{E}\left[\mathbf{X}_{j, d}^{T} \mathbb{E}[\boldsymbol{\eta}_{j, d}\boldsymbol{\eta}_{i, d}^{T}| \mathbf{X}_{i, d} ]\mathbf{X}_{i, d}\right]\right] \no\\
&&\overset{(a)}= \frac{1}{l^{2}} \sum_{i=1}^{l} \text{trace}\left[\mathbb{E}\left[\mathbf{X}_{i, d}^{T} \sigma^2\mathbf{I}\mathbf{X}_{i, d}\right]\right] \no\\
&&= \frac{1}{l^{2}} \sum_{i=1}^{l} \sigma^2\text{trace}\left[\mathbf{I}\right] = \frac{d^2}{t} \sigma^2,
\end{eqnarray}
in which (a) is true due to the fact that $\mathbf{X}_{k,d}$ is independent of $\boldsymbol{\eta}_{k, d}$. For the last item, we have
\begin{eqnarray}
&&\mathbb{E}[\boldsymbol{\theta^{*}}^{T}\mathbf{W}_{0}\mathbf{W}_{t}^{-2}\mathbf{X}_{t}^{T}\boldsymbol{\eta}_{t}] \no\\
&&= \boldsymbol{\theta^{*}}^{T}\mathbf{W}_{0}(\mathbf{W}_{0}+l\mathbf{I})^{-2}\mathbb{E}[\mathbf{X}_{t}^{T}\boldsymbol{\eta}_{t}] \no\\
&&= \boldsymbol{\theta^{*}}^{T}\mathbf{W}_{0}(\mathbf{W}_{0}+l\mathbf{I})^{-2}\mathbb{E}\left[\sum_{k=1}^{l} \mathbf{X}_{k,d}^{T}\boldsymbol{\eta}_{k, d}\right] = 0. \no
\end{eqnarray}
As a result,
\begin{eqnarray}
\mathbb{E}[||(\hat{\boldsymbol{\theta}}_{t} - \boldsymbol{\theta^{*}})||_{2}^2] &\leq&  \frac{1}{l^2}\boldsymbol{\theta^{*}}^{T}\mathbf{W}_{0}^2\boldsymbol{\theta^{*}} + \frac{d^2}{t} \sigma^2 \no\\
&=& \frac{d^2}{t^2}\boldsymbol{\theta^{*}}^{T}\mathbf{W}_{0}^2\boldsymbol{\theta^{*}} + \frac{d^2}{t} \sigma^2.
\end{eqnarray}
Therefore, the first part of Theorem \ref{thm:proposed_MSE} is established.

We then show the second part of Theorem \ref{thm:proposed_MSE}. To this end, we use the concentration inequality in Theorem 2.1 \cite{Hsu:ECP:12}. In particular, let $\mathbf{A} \in \mathbb{R}^{n\times n}$ be a square matrix and let $\mathbf{x} \in \mathbb{R}^{n\times1}$ be a sub-Gaussian random vector with zero mean and proxy $\sigma^2$. It has been shown that
\begin{eqnarray}&&||\mathbf{A}\mathbf{x}||_{2}^{2} \leq \nonumber\\
&&\sigma^2 \left( \text{trace}(\mathbf{A}^{T}\mathbf{A}) + 2\sqrt{\text{trace}((\mathbf{A}^{T}\mathbf{A})^2)t}+||\mathbf{A}||_{2}t \right)\nonumber\end{eqnarray}
holds with probability at least $1-e^{-t}$.  % paper ``A tail inequality for quadratic forms of subgaussian random vecorts'', $||\mathbf{A}||_{2}$ is the spectra norm of $\mathbf{A}$ ).

It is easy to find that
\begin{eqnarray}
&&\text{trace}(\mathbf{X}_{t}^{T}\mathbf{X}_{t}) = ld, \no\\
&&\text{trace}((\mathbf{X}_{t}^{T}\mathbf{X}_{t})^2) = l^2d, \no\\
&&||\mathbf{X}_{t}||_{2} = \lambda_{\max}(\mathbf{X}_{t}^{T}\mathbf{X}_{t})^{1/2}=l^{1/2}. \no
\end{eqnarray}
Hence, using above tail inequality, we have
\begin{eqnarray}
P\left(\boldsymbol{\eta}_{t}^{T}\mathbf{X}_{t}\mathbf{X}_{t}^{T}\boldsymbol{\eta}_{t} \leq \sigma^2( ld + 2\sqrt{l^2dt} + \sqrt{lt^2} )\right) \geq 1-e^{-t}.
\end{eqnarray}
Recall \eqref{eq:upperBound}, we have
\begin{eqnarray}
||\hat{\boldsymbol{\theta}}_{t} - \boldsymbol{\theta^{*}}||_{2}^{2}
&\leq& \frac{1}{l^2}\boldsymbol{\theta^{*}}^{T}\mathbf{W}_{0}\mathbf{W}_{0}\boldsymbol{\theta^{*}} + \frac{1}{l^2}\boldsymbol{\eta}_{t}^{T}\mathbf{X}_{t}\mathbf{X}_{t}^{T}\boldsymbol{\eta}_{t} \no \\
&+& 2|\boldsymbol{\theta}^{*T}\mathbf{W}_{0}\mathbf{W}_{t}^{-2}\mathbf{X}_{t}^{T}\boldsymbol{\eta}_{t}|. \no
\end{eqnarray}
For the last item on the right hand of the inequality, we have
\begin{eqnarray}
&&|\boldsymbol{\theta}^{*T}\mathbf{W}_{0}\mathbf{W}_{t}^{-2}\mathbf{X}_{t}^{T}\boldsymbol{\eta}_{t}| \no\\
&&\leq \lambda_{\max}(\mathbf{W}_{0})\lambda_{\max}(\mathbf{W}_{t}^{-2})|\boldsymbol{\theta}^{*T}\mathbf{X}_{t}^{T}\boldsymbol{\eta}_{t}| \no\\
&&=\frac{\lambda_{\max}(\mathbf{W}_{0})}{\lambda_{\min}(\mathbf{W}_{t})^2}|\boldsymbol{\eta}^{T}\mathbf{X}_{t}\boldsymbol{\theta}^{*}\boldsymbol{\theta}^{*T}\mathbf{X}_{t}^{T}\boldsymbol{\eta}_{t}|^{1/2}\no\\
&&\leq \frac{\lambda_{\max}(\mathbf{W}_{0})||\boldsymbol{\theta}^{*}||}{\lambda_{\min}(\mathbf{W}_{0}+l\mathbf{I})^2}|\boldsymbol{\eta}^{T}\mathbf{X}_{t}\mathbf{X}_{t}^{T}\boldsymbol{\eta}_{t}|^{1/2}\no\\
&&\leq \frac{\lambda_{\max}(\mathbf{W}_{0})||\boldsymbol{\theta}^{*}||}{l^2}(\boldsymbol{\eta}^{T}\mathbf{X}_{t}\mathbf{X}_{t}^{T}\boldsymbol{\eta}_{t})^{1/2}.\no
\end{eqnarray}
Therefore
\begin{eqnarray}
||\hat{\boldsymbol{\theta}}_{t} - \boldsymbol{\theta^{*}}||_{2}^{2}
&\leq& \frac{1}{l^2}\boldsymbol{\theta^{*}}^{T}\mathbf{W}_{0}\mathbf{W}_{0}\boldsymbol{\theta^{*}} + \frac{1}{l^2}\boldsymbol{\eta}_{t}^{T}\mathbf{X}_{t}\mathbf{X}_{t}^{T}\boldsymbol{\eta}_{t} \no \\
&+& \frac{\lambda_{\max}(\mathbf{W}_{0})||\boldsymbol{\theta}^{*}||}{l^2}(\boldsymbol{\eta}^{T}\mathbf{X}_{t}\mathbf{X}_{t}^{T}\boldsymbol{\eta}_{t})^{1/2}. \no
\end{eqnarray}
Then the event
\begin{eqnarray}
\left\{ \boldsymbol{\eta}_{t}^{T}\mathbf{X}_{t}\mathbf{X}_{t}^{T}\boldsymbol{\eta}_{t} \leq \sigma^2( ld + 2\sqrt{l^2dt} + \sqrt{lt^2} ) \right\} \no
\end{eqnarray}
holds with probability at least $1-e^{-t}$ indicates that the event
\begin{eqnarray}
\left\{||\hat{\boldsymbol{\theta}}_{t} - \boldsymbol{\theta^{*}}||_{2}^2 \leq \frac{1}{l^2}\boldsymbol{\theta^{*}}^{T}\mathbf{W}_{0}^2\boldsymbol{\theta^{*}} + \frac{\sigma^2}{l^2}( ld + 2\sqrt{l^2dt} + \sqrt{l}t) \right.\no\\
\left. + \lambda_{\max}(\mathbf{W}_{0})||\boldsymbol{\theta}^{*}||\frac{1}{l^2}( ld + 2\sqrt{l^2dt} + \sqrt{l}t)^{1/2} \right\}. \no
\end{eqnarray}
holds with probability at least $1-e^{-t}$. Since $t=ld$, we then can obtain
\begin{small}
\begin{eqnarray}
%&&P\left( ||\hat{\boldsymbol{\theta}}_{t} - \boldsymbol{\theta^{*}}||_{2}^2 \geq \frac{\boldsymbol{\theta^{*}}^{T}\mathbf{W}_{0}^2\boldsymbol{\theta^{*}}}{l^2} + \frac{\sigma^2( ld + 2\sqrt{l^2dt} + \sqrt{l}t)}{l^2} \right) \no\\
&&P\left( ||\hat{\boldsymbol{\theta}}_{t} - \boldsymbol{\theta^{*}}||_{2}^2 \geq \frac{d^2}{t^2}\boldsymbol{\theta^{*}}^{T}\mathbf{W}_{0}^2\boldsymbol{\theta^{*}} + \sigma^2\left(\frac{d^2}{t}+3 \sqrt{\frac{d^3}{t}} \right) \right.\no\\
&&\left.+ \lambda_{\max}(\mathbf{W}_{0})||\boldsymbol{\theta}^{*}||\frac{d\sigma^2}{t}\left(\frac{d^2}{t}+3 \sqrt{\frac{d^3}{t}}\right)^{1/2} \right) \no\\
&&= P\left( ||\hat{\boldsymbol{\theta}}_{t} - \boldsymbol{\theta^{*}}||_{2}^2 \geq \frac{3\sigma^2 d^{3/2}}{t^{1/2}} + O\left(\frac{\sigma^2 d^2}{t}\right) \right) \no\\
&&\leq e^{-t}.
\end{eqnarray}
\end{small}

\section{Proof of Theorem \ref{thm:lowerbound}} \label{adp:lowerbound}
For the estimator
$\hat{\boldsymbol{\theta}}_{t} = (\mathbf{X}_{t}^{T}\mathbf{X}_{t} + \mathbf{W}_{0})^{-1}\mathbf{X}_{t}^{T}\mathbf{Y}_{t}, $
we have
\begin{eqnarray}
\hat{\boldsymbol{\theta}}_{t} - \boldsymbol{\theta^{*}} = - \mathbf{W}_{t}^{-1}\mathbf{W}_{0}\boldsymbol{\theta^{*}}+ \mathbf{W}_{t}^{-1}\mathbf{X}_{t}^{T}\boldsymbol{\eta}_{t}. \no
\end{eqnarray}
Therefore
\begin{eqnarray}
&&(\hat{\boldsymbol{\theta}}_{t} - \boldsymbol{\theta^{*}})^{T}\mathbf{W}_{t}^{2} (\hat{\boldsymbol{\theta}}_{t} - \boldsymbol{\theta^{*}}) \no\\
&&= (-\boldsymbol{\theta^{*}}^{T}\mathbf{W}_{0}+\boldsymbol{\eta}_{t}^{T}\mathbf{X}_{t})(-\mathbf{W}_{0}\boldsymbol{\theta^{*}}+\mathbf{X}_{t}^{T}\boldsymbol{\eta}_{t}) \no\\
&&= \boldsymbol{\theta^{*}}^{T}\mathbf{W}_{0}^{2}\boldsymbol{\theta^{*}} - 2\boldsymbol{\theta^{*}}^{T}\mathbf{W}_{0}\mathbf{X}_{t}^{T}\boldsymbol{\eta}_{t} +  \boldsymbol{\eta}_{t}^{T}\mathbf{X}_{t}\mathbf{X}_{t}^{T}\boldsymbol{\eta}_{t}. \no
\end{eqnarray}
We then analyze the expectation of the three items on the right hand side of the equality one by one. We note that the first item $\boldsymbol{\theta^{*}}^{T}\mathbf{W}_{0}^{2}\boldsymbol{\theta^{*}}$ is a constant; for the second item, we have
\begin{eqnarray}
\mathbb{E}\left[\boldsymbol{\theta^{*}}^{T}\mathbf{W}_{0}\mathbf{X}_{t}^{T}\boldsymbol{\eta}_{t} \right] = \boldsymbol{\theta^{*}}^{T}\mathbf{W}_{0}\mathbb{E}\left[\sum_{i=1}^{t} \mathbf{x}_{i}\eta_{i} \right] = 0.
\end{eqnarray}
For the third item, we have
\begin{eqnarray}
&&\mathbb{E}[\boldsymbol{\eta}_{t}^{T}\mathbf{X}_{t}\mathbf{X}_{t}^{T}\boldsymbol{\eta}_{t}] \no\\
&&= \mathbb{E}\left[ \left(\sum_{i=1}^{t} \mathbf{x}_{i}\eta_{i} \right)^2 \right]\no\\
&&= \mathbb{E}\left[ \sum_{i=1}^{t}\sum_{j=1}^{t} \eta_{i}\mathbf{x}_{i}^{T} \mathbf{x}_{j}\eta_{j} \right] \no\\
&&= \mathbb{E}\left[ \sum_{i=1}^{t} \eta_{i}\mathbf{x}_{i}^{T} \mathbf{x}_{i}\eta_{i} + 2\sum_{i=1}^{t}\sum_{j=i+1}^{t} \eta_{i}\mathbf{x}_{i}^{T} \mathbf{x}_{j}\eta_{j} \right] \no\\
&&=  \sum_{i=1}^{t} \mathbb{E}\left[ \eta_{i}\mathbf{x}_{i}^{T} \mathbf{x}_{i}\eta_{i} \right] + 2\sum_{i=1}^{t}\sum_{j=i+1}^{t} \mathbb{E}\left[\eta_{i}\mathbf{x}_{i}^{T} \mathbf{x}_{j}\eta_{j} \right]. \no
\end{eqnarray}

Since $\eta_{i}$ is independent of $\mathbf{x}_{i}$, $\eta_{j}$ is independent of $\eta_{i}\mathbf{x}_{i}^{T} \mathbf{x}_{j}$ for $j>i$ and $||\mathbf{x}_{i}||^{2}=1$, we have
\begin{eqnarray}
&&\mathbb{E}\left[ \eta_{i}\mathbf{x}_{i}^{T} \mathbf{x}_{i}\eta_{i} \right] = \mathbb{E}\left[ \mathbf{x}_{i}^{T} \mathbf{x}_{i} \right] \mathbb{E}\left[ \eta_{i}^2 \right] = \sigma^2, \no\\
&&\mathbb{E}\left[\eta_{i}\mathbf{x}_{i}^{T} \mathbf{x}_{j}\eta_{j} \right] = \mathbb{E}\left[\eta_{i}\mathbf{x}_{i}^{T} \mathbf{x}_{j}\right] \mathbb{E}\left[ \eta_{j} \right] = 0. \no
\end{eqnarray}
Therefore
\begin{eqnarray}
\mathbb{E}[\boldsymbol{\eta}_{t}^{T}\mathbf{X}_{t}\mathbf{X}_{t}^{T}\boldsymbol{\eta}_{t}] =  \sum_{i=1}^{t} \sigma^2 = t\sigma^2.
\end{eqnarray}
As a result
$$\mathbb{E}[(\hat{\boldsymbol{\theta}}_{t} - \boldsymbol{\theta^{*}})^{T}\mathbf{W}_{t}^2 (\hat{\boldsymbol{\theta}}_{t} - \boldsymbol{\theta^{*}})]= \boldsymbol{\theta^{*}}^{T}\mathbf{W}_{0}^{2}\boldsymbol{\theta^{*}} + t\sigma^2.$$

Since $\mathbf{W}_{t} = \mathbf{W}_{0} + \mathbf{X}_{t}^{T}\mathbf{X}_{t}$, we have
\begin{eqnarray}
&&\lambda_{\max}(\mathbf{W}_{t}) \leq \text{trace}(\mathbf{W}_{t}) = \text{trace}(\mathbf{W}_{0}) + \sum_{i=1}^{t} \mathbf{x}_{i}^{T}\mathbf{x}_{i} \no\\
&&\hspace{17mm}= \text{trace}(\mathbf{W}_{0}) + t, \no\\
&&\lambda_{\max}(\mathbf{W}_{t}^2) = \lambda_{\max}^{2}(\mathbf{W}_{t}) \leq (\text{trace}(\mathbf{W}_{0}) + t)^2. \no
\end{eqnarray}

Therefore,
\begin{eqnarray}
&&\hspace{-8mm}\mathbb{E}[|| \hat{\boldsymbol{\theta}}_{t} - \boldsymbol{\theta^{*}} ||_{2}^2] \no\\
&&\hspace{-8mm}\geq \frac{1}{\lambda_{\max}(\mathbf{W}_{t}^2)} \mathbb{E}[(\hat{\boldsymbol{\theta}}_{t} - \boldsymbol{\theta^{*}})^{T}\mathbf{W}_{t}^2 (\hat{\boldsymbol{\theta}}_{t} - \boldsymbol{\theta^{*}})] \no\\
%&&\hspace{-8mm}\geq \frac{\mathbb{E}\left[ \boldsymbol{\theta^{*}}^{T}\mathbf{W}_{0}^{2}\boldsymbol{\theta^{*}}\right] - \mathbb{E}\left[2\boldsymbol{\theta^{*}}^{T}\mathbf{W}_{0}\mathbf{X}_{t}^{T}\boldsymbol{\eta}_{t} \right] + \mathbb{E}\left[ \boldsymbol{\eta}_{t}^{T} \mathbf{X}_{t}\mathbf{X}_{t}^{T}\boldsymbol{\eta}_{t} \right]}{(\text{trace}(\mathbf{W}_{0}) + t)^{2}} \no\\
&&\hspace{-8mm}\geq \frac{\boldsymbol{\theta^{*}}^{T}\mathbf{W}_{0}^{2}\boldsymbol{\theta^{*}} + t\sigma^2}{(\text{trace}(\mathbf{W}_{0}) + t)^{2}},
\end{eqnarray}
in which the first inequality is because of
$$(\hat{\boldsymbol{\theta}}_{t} - \boldsymbol{\theta^{*}})^{T}\mathbf{W}^{2}_{t}(\hat{\boldsymbol{\theta}}_{t} - \boldsymbol{\theta^{*}}) \leq \lambda_{\max}(\mathbf{W}_{t}^2)||(\hat{\boldsymbol{\theta}}_{t} - \boldsymbol{\theta^{*}})||_{2}^{2}. $$
Then, the result of Theorem \ref{thm:lowerbound} can be obtained by taking $t\rightarrow\infty$.
\end{document}